\pdfoutput=1

\documentclass[11pt]{article}

\usepackage{EACL2023}

\usepackage{times}
\usepackage{latexsym}

\usepackage[T1]{fontenc}

\usepackage[utf8]{inputenc}

\usepackage{microtype}

\usepackage{inconsolata}

\usepackage[hang,flushmargin]{footmisc}  
\usepackage{amsfonts,amsmath,amssymb}
\usepackage{bold-extra}   
\usepackage{subcaption}
\usepackage{graphicx}
\usepackage{multirow}
\usepackage{enumitem}
\usepackage{url}

\usepackage{amsmath}
\usepackage[linesnumbered,ruled]{algorithm2e}
\usepackage{pgfplots}
\pgfplotsset{compat=1.9}
\usepgfplotslibrary{colorbrewer}
\usepackage{tikz}
\usetikzlibrary{positioning}

\newcommand{\squishlist}{
 \begin{list}{$\bullet$}
  { \setlength{\itemsep}{0pt}
     \setlength{\parsep}{3pt}
     \setlength{\topsep}{3pt}
     \setlength{\partopsep}{0pt}
     \setlength{\leftmargin}{1.5em}
     \setlength{\labelwidth}{1em}
     \setlength{\labelsep}{0.5em} } }

\newcommand{\squishlisttwo}{
 \begin{list}{$\bullet$}
  { \setlength{\itemsep}{0pt}
     \setlength{\parsep}{0pt}
    \setlength{\topsep}{0pt}
    \setlength{\partopsep}{0pt}
\setlength{\leftmargin}{2em}
\setlength{\labelwidth}{1.5em}
\setlength{\labelsep}{0.5em} } }

\newcommand{\squishend}{
\end{list}  }

\title{Closed-book Question Generation via Contrastive Learning}

\author{Xiangjue Dong$^1$, Jiaying Lu$^2$\thanks{\hspace{0.2cm}Equal Contribution}, Jianling Wang$^1$\footnotemark[1], James Caverlee$^1$\\
        $^1$ Texas A\&M University, $^2$ Emory University \\ \small\texttt{\{xj.dong, jlwang, caverlee\}@tamu.edu, jiaying.lu@emory.edu}}

\begin{document}
\maketitle
\begin{abstract}

Question Generation (QG) is a fundamental NLP task for many downstream applications. Recent studies on \textit{open-book} QG, where supportive answer-context pairs are provided to models, have achieved promising progress.
However, generating natural questions under a more practical \textit{closed-book} setting that lacks these supporting documents still remains a challenge.
In this work, we propose a new QG model for this closed-book setting that is designed to better understand the semantics of long-form abstractive answers and store more information in its parameters through contrastive learning and an answer reconstruction module.
Through experiments, we validate the proposed QG model on both public datasets and a new WikiCQA dataset. Empirical results show that the proposed QG model outperforms baselines in both automatic evaluation and human evaluation.
In addition, we show how to leverage the proposed model to improve existing question-answering systems. These results further indicate the effectiveness of our QG model for enhancing \textit{closed-book} question-answering tasks.



 


\end{abstract}

\section{Introduction}

Question Generation (QG) has a wide range of applications, such as generating questions for exams~\cite{jia-etal-2021-eqg, lelkes2021quiz, dugan2022feasibility} or children's story books~\cite{zhao-etal-2022-educational, Yao2022ItIA}, recommending questions for users in a dialogue system~\cite{shukla-etal-2019-ask, laban-etal-2020-whats}, improving visual~\cite{Li_2018_CVPR,lu2022good} or textual question-answering tasks~\cite{duan-etal-2017-question, lewis2019unsupervised, zhang2019addressing, sultan-etal-2020-importance, lyu-etal-2021-improving}, asking clarification questions~\cite{rao-daume-iii-2019-answer, yu-etal-2020-interactive, ren-etal-2021}, and generating queries for SQL~\cite{wu2021data} or multimodal documents~\cite{kim-etal-2021-query}.

\noindent Previous works on QG are mainly under the \textit{open-book} setting, which aims to generate questions based on factoid or human-generated short answers under the assumption that there is access to external knowledge like retrieved documents or passages~\cite{du-etal-2017-learning, zhao-etal-2018-paragraph, kim2019improving, fei-etal-2021-iterative}. After~\citet{roberts-etal-2020-much} demonstrated that feeding a large pre-trained model input questions alone \textit{without} any external knowledge can lead to competitive results with retrieval-based methods on open-domain question-answering benchmarks, there is an increasing interest in the \textit{closed-book} setting. This closed-book setting is appealing in practice and can be widely applied, e.g., in question suggestion~\cite{laban-etal-2020-whats, yin2021summary},  query recommendation~\cite{kim-etal-2021-query}, and other practical settings where extensive external knowledge is unavailable. 

However, generating questions without access to such external knowledge is challenging for two key reasons. First, without access to retrieved documents (or passages), simple open-domain strategies like basing the answers on these documents (or passages) are not possible under the closed-book setting. Instead, models must rely on the answers alone. Second, the data used by most of the closed-book works~\cite{lewis-etal-2021-paq, wang-etal-2021-generative} are variants of existing open-domain datasets, e.g., SQuAD~\cite{rajpurkar-etal-2018-know}, TriviaQA~\cite{joshi-etal-2017-triviaqa}, WebQuestions~\cite{berant-etal-2013-semantic} that ignore the answer-related passages. These answers in open-book works are usually short, e.g., entities, and easier to be remembered by the language model and stored in the parameters of the model than long-form answers. Thus, this leads to our motivating research question -- \textit{How can we empower a QG model to better understand the semantics of long-form abstractive answers and store more information in its parameters?}






To tackle the aforementioned challenges existing in the closed-book setting, this paper proposes a new QG model with two unique characteristics: (i) a contrastive learning loss designed to better understand the semantics of the answers and the semantic relationship between answers and ground-truth questions at a contextual-level; and (ii) an answer reconstruction loss designed to measure the answerability of the generated question. Contrastive learning has shown promising results in many NLP tasks, e.g., ~\cite{giorgi-etal-2021-declutr, gao-etal-2021-simcse, yang-etal-2021-contrastive-representation} and aligns positive pairs better with available supervised signals~\cite{gao-etal-2021-simcse}; here we show how to learn question representations by distinguishing features of correct question-answer pairs from features of incorrectly linked question-answer pairs. Further, to ensure the generated questions are of good quality and can be answered by the answer that is used for question generation, we frame the model as a generation-reconstruction process~\cite{cao-etal-2019-semantic, zhu2020dual}, by predicting the original answers given the generated questions by a pre-trained seq2seq model.
In addition, we introduce a new closed-book dataset with long-form abstractive answers -- WikiCQA -- to complement existing datasets like GooAQ~\cite{khashabi-etal-2021-gooaq-open} and ELI5~\cite{fan-etal-2019-eli5} and show how to leverage our model to generate synthetic data to improve closed-book question-answering tasks.

Through experiments, we find that the proposed QG model shows improvement through both automatic and human evaluation metrics on WikiCQA and two public datasets. Compared to the baseline, the proposed QG framework shows an improvement of up to 2.0\%, 2.7\%, and 1.8\% on the ROUGE-L score on WikiCQA, GooAQ-S, and ELI5, respectively, and 1.3\% and 2.6\% in terms of relevance and correctness. Furthermore, we leverage the QG framework to generate synthetic QA data from WikiHow summary data and pre-train a closed-book QA model on it in both an unsupervised and semi-supervised setting. The performance is evaluated on both seen (WikiCQA) and unseen (GooAQ-S, ELI5) datasets. We find consistent improvements across these datasets, indicating the QG model's effectiveness in enhancing closed-book question-answering tasks. 

In conclusion, our contributions can be summarized as follows:
\squishlist
\item We propose a contrastive QG model, which to our knowledge is the first work to explore contrastive learning for QG under a closed-book setting.
\item The proposed model outperforms baselines on three datasets. The human evaluation also indicates that the questions generated by our model are more informative compared to other baselines.
\item We leverage the QG model as a data augmentation strategy to generate large-scale QA pairs. Consistent improvements shown on both seen datasets and unseen datasets indicate the QG model's effectiveness in enhancing closed-book question-answering tasks.
\squishend

\section{Related Work}

Many previous works on QG are under the open-book setting, which takes factoid short answers~\cite{rajpurkar-etal-2016-squad} or human-generated short answers~\cite{kocisky-etal-2018-narrativeqa} with the corresponding passages to generate questions~\cite{zhang-etal-2021-review}.
Early approaches for question generation rely on rule-based methods~\cite{labutov-etal-2015-deep, khullar-etal-2018-automatic}. To bypass hand-crafted rules and sophisticated pipelines in QG, \citet{du-etal-2017-learning} introduce a vanilla RNN-based sequence-to-sequence approach with an attention mechanism. The recently proposed pre-trained transformer-based frameworks~\cite{lewis-etal-2020-bart,raffel-etal-2020-exploring} also improve the performance of QG. 
In addition, ~\citet{sultan-etal-2020-importance} shows that the lexical and factual diversity of QG provides better QA training.
However, their success can not directly adapt to the closed-book setting, where the model is supposed to generate questions solely relying on answers.
In this work, we explore the widely applicable closed-book QG setting, which is still under-explored.






\smallskip
\noindent\textbf{Contrastive Learning} aims to pull semantically similar neighbors close and push non-neighbors apart. It has achieved great success under both supervised and unsupervised settings.  
In pioneer works, the contrastive loss function \cite{hadsell2006dimensionality,chopra2005learning} has been proposed as a training objective in deep metric learning considering both similar and dissimilar pairs. Recently,  \citet{chen-etal-2020-simple} proposes the SimCLR framework to learn useful visual representations. Viewing contrastive learning as dictionary look-up, \citet{he-etal-2020-momentum} present Momentum Contrast (MoCo) to build dynamic dictionaries for contrastive learning. 
Some works apply contrastive learning into the NLP domain to learn better sentence representations~\cite{giorgi-etal-2021-declutr, gao-etal-2021-simcse}.
In addition, contrastive learning has been applied in multilingual neural machine translation~\cite{pan-etal-2021-contrastive}, abstractive summarization~\cite{liu-liu-2021-simcls}, and multi-document question generation~\cite{cho-etal-2021-contrastive}.
The recent most relevant work is~\cite{yang-etal-2021-contrastive-representation}, where they design two contrastive losses for paraphrase generation.
In this work, we adopt contrastive learning for improving representation learning in question generation under a closed-book setting. 

\section{Proposed Approach}
\label{sec:approach}

\begin{figure}[htbp!]
\centering
\includegraphics[width=\linewidth]{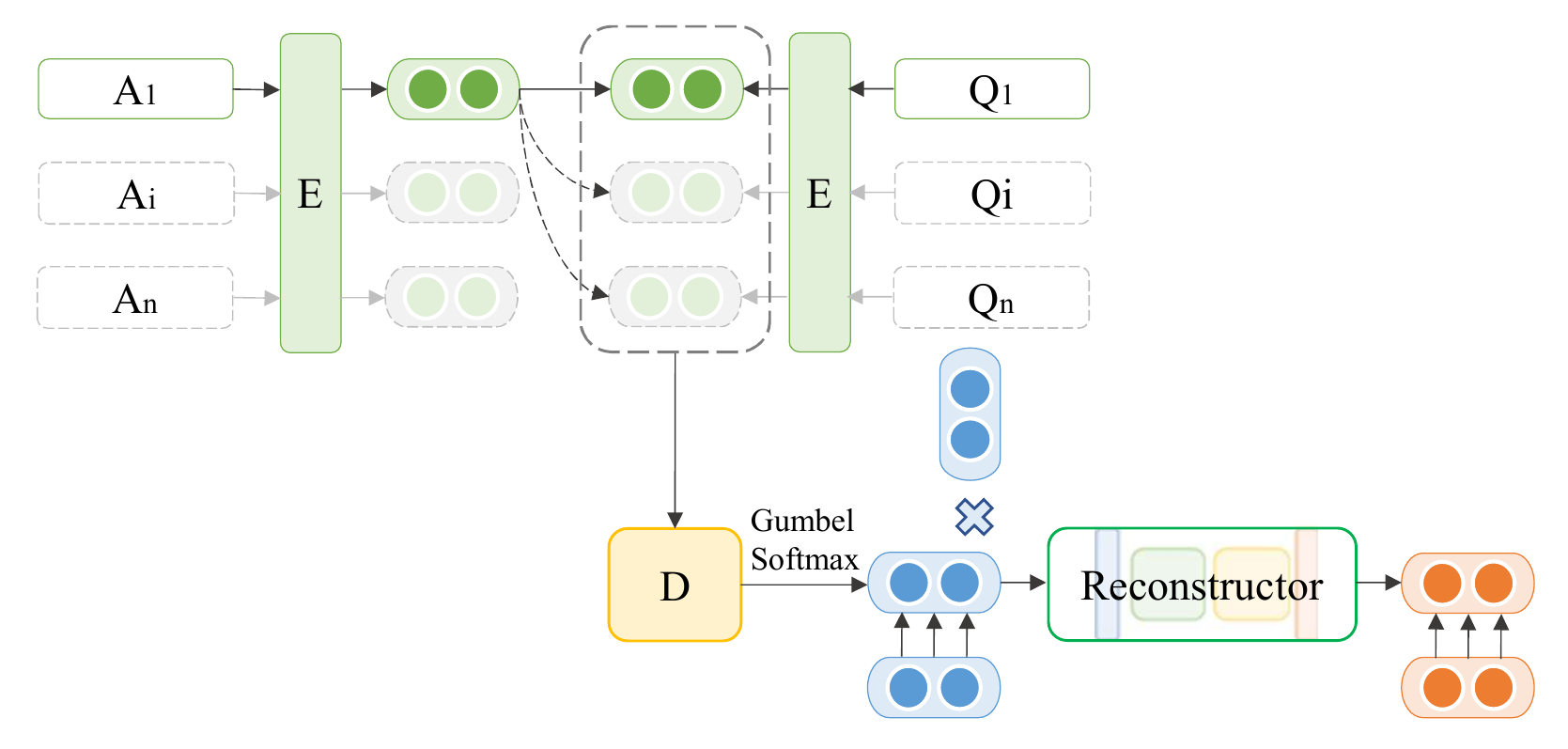}
\caption{An overview of the proposed closed-book QG framework, which consists of three parts: contrastive learning, question generation, and answer construction. \textbf{$A_{i}$}: represents answer i; \textbf{$Q_{i}$} represents question i.}    
\label{fig:qg-framework}
\end{figure}

To answer our research question -- How can we empower a QG model to better understand the semantics of long-form abstractive answers and store more information in its parameters? -- we propose a closed-book QG model, which generates questions directly without access to external knowledge.
Formally, given an answer sentence $x$, a closed-book QG engine generates a natural question $y$.
Figure~\ref{fig:qg-framework} illustrates an overview of the proposed QG framework, which consists of three parts: question generation, contrastive learning, and answer reconstruction.
The framework is optimized with the joint losses from these three parts simultaneously.

\vspace{-0.1cm}

\subsection{Question Generation}
\label{subsec:qg}
We first focus on question generation through a  sequence-to-sequence architecture which consists of an encoder and a decoder ~\cite{sutskever-etal-2014-sequencetosequence,vaswani-etal-2017-attention}. 
The encoder takes an input sequence of source words $\textbf{x} = (x_1, x_2, \dots, x_n)$ and maps it to a sequence of continuous representations $\textbf{z} = (z_1, z_2, \dots, z_n)$. 
Then, the decoder takes $\textbf{z}$ and generates a sequence of target words $\textbf{y} = (y_1, y_2, \dots, y_m)$ at a time.
The closed-book QG task is defined as finding $\hat{\textbf{y}}$:
\begin{equation}
    \hat{\textbf{y}}=\underset{\textbf{y}}{\arg\max} P(\textbf{y}|\textbf{x}),
\end{equation}
where $P(\textbf{y}|\textbf{x})$ is the conditional likelihood of the predicted question sequence $\textbf{y}$ given answer $\textbf{x}$.
\begin{equation}
    P(\textbf{y}|\textbf{x})=\prod_{i=1}^{T}p(y_{t}|\textbf{y}_{<t},\textbf{x}),
\end{equation}

Given the answer-question pairs, the training objective of the generation part in the proposed framework is to minimize the Negative Log-Likelihood (NLL) of the training data,
\begin{equation}\label{eq:qg}
    \mathcal{L}_{qg}=-\sum_{i=1}^{N}\log p(q_{i}|A),
\end{equation}
where $q_{i}$ is the $i$-th token in the generated question and $A$ is the answer.

A naive question generation model will generate questions based on answers but lacks a rich model of the semantics of answers nor can it guarantee the generated questions have a semantic relationship with the answers. Intuitively, an encoded answer should be similar to its question and dissimilar to others. In addition, the generated question should be able to be answered by the answers. Hence, this motivates the following contrastive learning and answer reconstruction modules.

\subsection{Contrastive Learning}
\label{subsec:cl}

Contrastive learning aims to pull positive pairs and push apart negative pairs to learn effective representations. Further, the supervised signals can produce better sentence embeddings by improving alignment between positive pairs~\cite{chen-etal-2020-simple}. An effective QG model should be able to understand the semantics of the answers and the semantic relationship with the ground-truth questions. Especially, the encoded answer should have semantic similarity with its ground-truth question and dissimilarity with other questions. Thus, aiming to learn a similarity function that pulls the distance between the answer sequence representation and its ground-truth question sequence representation closer, we design a contrastive loss in the representation space. Specifically, given a positive pair $\mathcal{S}=\left\{(\textbf{x}_{i}, \textbf{y}_{i})  \right\}_{i=1}^{n}$, where $\textbf{x}_{i}$ and $\textbf{y}_{i}$ are semantically related inputs, the other $2(n-1)$ examples within a mini-batch are treated as negative examples. The training objective for $(\textbf{x}_{i}, \textbf{y}_{i})$ is:
\begin{equation}\label{eq:cl}
    \mathcal{L}_{cl}=-\log\frac{\exp(\text{sim}(\textbf{z}_{x_i},\textbf{z}_{y_i})/\tau_{cl})}{\sum_{i=1}^{2n}\exp(\text{sim}(\textbf{z}_{x_i},\textbf{z}_{y_i})/\tau_{cl})},
\end{equation}
where $\textbf{z}_{\textbf{x}_i}$ and $\textbf{z}_{\textbf{y}_i}$ is the representation of input $x_{i}$ and $y_{i}$, $\text{sim}(\textbf{z}_{i},\textbf{z}_{j})=\textbf{z}_{i}^{\top}\textbf{z}_{j}/\left\| \textbf{z}_{i} \right\|\left\| \textbf{z}_{j} \right\|$ denotes cosine similarity, and $\tau_{cl}$ is a temperature parameter.

In this work, aiming to learn better answer representations and force the encoder to drive representations of correct question-answer pairs closer than representations of incorrect question-answer pairs, we take the ground-truth question as the positive instance of an answer and fine-tune the model parameters based on the contrastive loss function (Eq.~\ref{eq:cl}), where $\textbf{z}$ is the embedding of the special token \texttt{[CLS]} from the transformer encoder, representing the meaning of the entire sentence.

\subsection{Answer Reconstruction}
\label{subsec:ar}

The questions that are generated by the model should be of good quality and should also be able to be answered by the answer that is used for question generation. To measure the answerability of the generated question, we design an answer reconstruction module, which uses a pre-trained seq2seq model to predict the original answer given the generated question. The loss is calculated by a negative log-likelihood loss function: 
\begin{equation}\label{eq:ar}
    \mathcal{L}_{ar}=-\sum_{i=1}^{N}\log p(a_{i}|Q)
\end{equation}
where $a_{i}$ is the $i$-th token in the answer and Q is the generated question.

A major challenge is that the generated questions from Section~\ref{subsec:qg} are not differentiable.
Gradients cannot be back-propagated directly.
To solve this challenge, we employ the Straight-Through (ST) Gumbel-Softmax for gradient computation~\cite{jang-etal-2017-categorical}.
The ST Gumbel-Softmax is a discrete version of the Gumbel-Softmax and takes different forward and backward paths. In the forward pass, the embedding is discretized by using the argmax, whereas in the backward pass the gradients are computed by the gumbel-softmax~\cite{qader-etal-2019-semi,lu2021weakly}. 
\begin{equation}
    y_{i} = \frac{\exp((\log(p_{i})+g_{i})/\tau_{gs})}{\sum_{j=1}^{|V|}\exp((\log(p_{j})+g_{j})/\tau_{gs})},
\end{equation}
where $\tau_{gs}$ is a temperature parameter and $g_{i}$ is the Gumbel noise drawn from a uniform distribution $(0,1)$. 
In this work, the one-hot embedding from ST Gumbel-Softmax is multiplied with the vocabulary embedding and then fed into the encoder of the pre-trained seq2seq model as the representation of the generated question.

\subsection{Overall Loss Function}
As a result, all three losses are summed together to provide the overall loss function $\mathcal{L}$ as follows:
\begin{equation}\label{eq:total}
    \mathcal{L} = \lambda_{1} \mathcal{L}_{qg} + \lambda_{2}\mathcal{L}_{cl} + \lambda_{3}\mathcal{L}_{ar},
\end{equation}
The weights $\lambda_{1}$,$\lambda_{2}$, and $\lambda_{3}$ are tuneable hyper-parameters to balance losses and the final objective is to minimize the overall loss.

The overall structure of the proposed QG framework is presented in Algorithm~\ref{alg:qg}. 

\begin{algorithm}[htbp!]
\caption{QG framework.}
\label{alg:qg}
\SetKwInput{KwData}{Input}
\SetKwInput{KwResult}{Output}
\KwData{Pre-trained language model $p(q|a)$, answer reconstruction model $p(a|q)$ and answer-question pairs}
\KwResult{Question generator $p(q|a)$}
\For{$i\gets1$ \KwTo $Epoch$}{
    $\hat{q} = p(q|a)$ \\
    Compute $\mathcal{L}_{qg}$ via Eq.~\ref{eq:qg} \\
    \tcc{contrastive learning}
    $a_i = \text{Encoder}(\texttt{[CLS]}\oplus a)[:,0:]$ \\
    $a^{+}_{i} = \text{Encoder}(\texttt{[CLS]}\oplus q)[:,0:]$ \\
    Get $\mathcal{L}_{cl}$ to $(a_i,a^{+}_{i})$ via Eq.~\ref{eq:cl} \\
    \tcc{answer reconstruction}
    $\hat{a} = p(a|\hat{q})$ \\
    Compute $\mathcal{L}_{ar}$ via Eq.~\ref{eq:ar} \\
    Calculate total loss $\mathcal{L}$ via Eq.~\ref{eq:total} \\
    Update generator $p(q|a)$ with $\mathcal{L}$}
\Return question generator $p(q|a)$
\end{algorithm}
\vspace{-0.3cm}
\section{Experimental Setup}

To evaluate the effectiveness of the proposed QG framework, we aim to answer the following research questions (RQ) via experiments: \textbf{RQ1}: Can this proposed QG framework improve the performance of the closed-book QG tasks? \textbf{RQ2}: Are the generated questions of good quality? That is, are they fluent and relevant to the answer? \textbf{RQ3}: Can this QG framework be leveraged as a good resource to generate synthetic data for the QA task? \textbf{RQ4}: How much does each component in the framework contribute?


\subsection{Dataset}
We conduct the experiments on two public datasets -- GooAQ-S~\cite{khashabi-etal-2021-gooaq-open} and ELI5~\cite{fan-etal-2019-eli5} -- and a new dataset we curate called WikiCQA.

\smallskip
\noindent \textbf{GooAQ-S} is a sub-sampled dataset from GooAQ which contains three different sub-tasks: short, snippet (i.e., multi-sentence description), and collection response questions~\cite{khashabi-etal-2021-gooaq-open}. 
Under the long-form closed-book setting, we adopt the snippet part (i.e., questions with snippet answers) for our experiments as in the original paper.
Furthermore, from the paper, the model performance on the snippet task does not vary much when supervised with 200K and 2M training instances. 
Thus, to improve the experimental efficiency, we take 200k instances from the snippet set and split them based on their original scripts.\footnote{\url{https://github.com/allenai/gooaq/blob/main/experiments/create\_splits.py}}

\smallskip
\noindent \textbf{ELI5}~\cite{fan-etal-2019-eli5} is a widely-used large-scale corpus for long-form question-answering with supporting web documents.
In this work, we use the question-answer pairs from the dataset and ignore the supporting context to fit the closed-book setting like~\cite{khashabi-etal-2021-gooaq-open}.
We follow the data split from huggingface.\footnote{\url{https://huggingface.co/datasets/eli5}}

\smallskip
\noindent \textbf{WikiCQA} is a new closed-book long-form QA dataset (Section~\ref{sec:dataset}) introduced here. It contains 20,202 question-answer pairs and is collected from a wiki-style website to complement existing datasets. 
We shuffle the data and split it to train, dev, and test sets by the ratio of 80\%/10\%/10\%. Table~\ref{tab:data-split} shows detailed data statistics and splits of these three datasets.
\begin{table}[htbp!]
\centering \small
\begin{tabular}{lrrr}
\hline
\textbf{Datasets} & \textbf{Train} & \textbf{Val} & \textbf{Test}\\
\hline
WikiCQA & 16,162 & 2,020 & 2,020 \\
GooAQ-S & 200,000 & 500 & 498 \\
ELI5 & 272,634 & 9,812 & 24,512 \\
\hline
\end{tabular}
\caption{Dataset statistics.}
\label{tab:data-split}
\end{table}

\subsection{WikiCQA}
\label{sec:dataset}

WikiCQA contains real user QA pairs collected from WikiHow\footnote{\url{https://www.wikihow.com}, under an Attribution-Noncommercial-Share Alike 3.0 Creative Commons License.}. WikiHow is a wiki-style website featuring over 200K how-to articles.  Different from the existing dataset, ELI5, containing questions/answers from the Reddit forum and leveraging evidence queried from the web to help answer the question~\cite{fan-etal-2019-eli5} and GooAQ, containing questions from search auto-complete and answers from answer boxes~\cite{khashabi-etal-2021-gooaq-open}, WikiCQA involves long-form question-answering grounded on WikiHow articles. 


\smallskip
\noindent \textbf{Dataset Construction} 
We construct the new dataset by collecting the question-answer pairs from the Q\&A section of articles on WikiHow.  
The questions and answers are related to the specific articles, asked by WikiHow users, and answered by a knowledgeable reader or editor\footnote{\url{https://www.wikihow.com/Use-wikiHow\#Reading-and-Learning-from-wikiHow}}. 
The questions can not be answered directly by the content of the article.
After removing duplicates and question-answer pairs with meaningless answers, 23,037 QA pairs remain. 
To keep the same format as other existing QA datasets (e.g., ELI5, GooAQ), we discard questions not starting with a question-type word (e.g., what, how). 
After the dataset processing steps, we arrive at 20,202 question-answer pairs. More details can be found in~\ref{sec:data-preprocess}.

In Table~\ref{tab:an-example}, we show an example article from Wikihow, which includes the title and summary of an article and question-answer pairs from the corresponding Q\&A section. From the example, we can see that answers are abstractive and long-form, written by real users, and not contained within the context passage. These question-answer pairs from the Q\&A section are collected in the newly constructed dataset. Thus, this dataset is different from reading comprehension datasets, where the answers are a short text span in context. More comparisons with ELI5 and GooAQ are shown in~\ref{sec:dataset-analysis}.

\begin{table}[htbp!]
\centering\small
\begin{tabular}{p{0.45\textwidth}}
\hline
\textbf{Title:} How to Prepare a Healthy Meal for Your Pet Dog. \\
\textbf{Summary:} To prepare a healthy meal for your dog, choose lean meat with the bones and fat removed, like chicken or beef \dots \\ \hline
\textbf{Q:} What can I feed my dog if I have run out of dog food? \\
\textbf{A:} In the short term, any bland human food such as chicken or white fish with rice or pasta is just fine \dots \\
\textbf{Q:} How much homemade dog food do you feed your dog? \\
\textbf{A:} Great question because it highlights one of the problems of feeding home prepared foods \dots \\
\textbf{Q:} What should I not feed my dog?  \\
\textbf{A:} There are many human foods that are toxic to dogs. Top of the list of foods NOT to give are \dots \\

\hline
\end{tabular}
\caption{Question-answer pairs from WikiHow Q\&A section. \textbf{Q}: question; \textbf{A}: answer.}
\label{tab:an-example}
\end{table}
\vspace{-0.3cm}

\subsection{Models \& Hyper-parameters}
We use BART-base~\cite{lewis-etal-2020-bart}, a widely used sequence-to-sequence framework with 12 layers and a hidden size of 1024, as the backbone model. Following previous works~\cite{fan-etal-2019-eli5,khashabi-etal-2021-gooaq-open}, we finetune it using answers as inputs and questions as outputs as the baselines. To get the answer reconstruction loss (Section~\ref{subsec:ar}), we use a BART-base model which is fine-tuned on the target QA datasets. 

Based on the dataset analysis in~\ref{sec:dataset-analysis}, we set the maximum sequence length to 128 for the question and 256 for the answer sentence to improve calculation efficiency.
We train the models for five epochs with a learning rate of $5\times 10^{-5}$ and evaluate the checkpoint for each epoch. 
We select the checkpoint with the highest ROUGE-L score on the validation set and report its corresponding score on the test set.
We run each model three times and record the average scores. 
After performing manual hyper-parameter searching, we set loss parameters $\lambda_{1}$, $\lambda_{2}$, and $\lambda_{3}$ in Equation~\ref{eq:total} to 1.0, 0.1, 0.1, respectively, which give the best ROUGE-L score on validation set. The temperature of contrastive loss $\tau_{cl}$ is set to 0.3. The experiments are run on 1 NVIDIA Tesla V100 GPU. The training time takes about 48 hours.

\subsection{Evaluation Metrics}
To evaluate the performance of the models, we use \textbf{ROUGE}~\cite{lin-2004-rouge} scores, which evaluate the $n$-grams recall of generated sentences with reference sentences, as automatic evaluation metrics~\cite{fan-etal-2019-eli5,khashabi-etal-2021-gooaq-open}. We report the F1 for ROUGE-1, ROUGE-2, ROUGE-L, and ROUGE-Lsum. ROUGE-1, 2, L measures the unigram, bigram, and longest common subsequence between the pair of sentences, respectively. The difference between ROUGE-L and ROUGE-Lsum is that ROUGE-Lsum splits text using `` \textbackslash n''. The higher ROUGE score indicates higher similarity between generated questions and references. 

We further perform human evaluations for the quality of generated questions in terms of \textbf{Fluency}: whether the questions are grammatically correct and fluent; \textbf{Relevance}: whether the questions are related to the answers; \textbf{Correctness}: whether the questions can be answered by the answers.

\section{Experiment Results}
In this section, we answer the four experimental research questions in turn.
\subsection{Generation Performance (RQ1)}
First, does the proposed QG framework have good performance? Table~\ref{tab:qg-results} shows the performance of the proposed closed-book QG model (denoted as \textbf{QG}$_{ours}$) on the new dataset and two public datasets compared with the baseline model trained only with question generation loss (denoted as \textbf{QG}$_{b}$). 
We observe that the proposed framework outperforms the baseline on three datasets for all ROUGE scores. For example, the performance increases by up to 2.0\%, 2.7\%, and 1.8\% on the ROUGE-L score, respectively, which means the proposed framework can generate questions having a longer longest common subsequence (LCS) with ground-truth questions. We attribute these improvements to how contrastive learning pulls the answer representations closer to the ground-truth questions and thus, the generated questions have a higher chance to have similar words, phrases, or sentences with the ground-truth questions. The results demonstrate the effectiveness of our proposed QG framework.

\begin{table}[htbp!]
\centering\small
\begin{tabular}{p{1cm}p{0.8cm}cccc}
\hline
\textbf{Dataset} & \textbf{Model} & \textbf{R-1} & \textbf{R-2} & \textbf{R-L} & \textbf{R-Lsum} \\
\hline
\multirow{2}{*}{WikiCQA} & QG$_{b}$ & 48.39 & 26.84 & 46.08 & 46.16  \\
                         & QG$_{ours}$ & \textbf{49.22} & \textbf{27.79} & \textbf{46.98} & \textbf{47.08} \\\hline
\multirow{2}{*}{GooAQ-S} & QG$_{b}$ & 44.26 & 19.73 & 41.11 & 41.08 \\
                       & QG$_{ours}$ & \textbf{45.26} & \textbf{20.69} & \textbf{42.20} & \textbf{42.06} \\\hline
\multirow{2}{*}{ELI5} & QG$_{b}$ & 28.62 & 10.10 & 25.93 & 26.23 \\
                      & QG$_{ours}$ & \textbf{29.15} & \textbf{10.36} & \textbf{26.40} & \textbf{26.69} \\
\hline
\end{tabular}
\caption{QG results on three datasets. \textbf{QG$_{b}$}: baseline model; \textbf{QG$_{ours}$}: our proposed QG framework.}
\label{tab:qg-results}
\end{table}
\subsection{Quality of Generated Questions (RQ2)}
Next, are the generated questions of good quality? To answer this question, we perform human evaluation to measure the quality of generated questions in terms of \textit{fluency}, \textit{relevance}, and \textit{correctness}.
We randomly sample 100 answer-question pairs from the WikiCQA dataset, which contains answers, ground-truth questions, and generated questions from the baseline model and our best model. 
Then, we ask three annotators to rate the generated question pairs, comparing them with the ground-truth questions.
The generated questions are rated on a 1-5 scale (5 for the best) from the aspects of the three aspects above.
Further, we calculate the percentage of questions that have higher ratings in each group.

\noindent Table~\ref{tab:human-evaluation} shows the average scores and the percentages of preferred questions on the three criteria.
Both models are equally good at generating fluent questions.
In terms of relevance and correctness, our approach shows 1.3\% and 2.6\% higher scores than the baseline method. These are consistent with what we expect: the contrastive learning part pulls the answer representation closer to the ground-truth question representation and generates more relevant questions; the reconstruction part can ensure that generated questions are of good quality and answerable.
In addition, among 100 answer-question pairs, 24.7\% and 23.3\% of generated questions from our best model are rated higher while 56.3\% and 57.4\% of them have the same rating as those from baselines, which indicates that there is still substantial room to improve.
To get a sense of the stability of the human evaluation results, we measure the inter-agreement among annotators using the Correlation Coefficient, which is 0.998, 0.944, and 0.976 in terms of these three aspects, showing excellent reliability.

\begin{table}[htbp!]
\centering\small
\begin{tabular}{ccccccc}
\hline
\multirow{2}{*}{\textbf{Model}} & \multicolumn{2}{l}{\textbf{fluency}} & \multicolumn{2}{l}{\textbf{relevance}} & \multicolumn{2}{l}{\textbf{correctness}}  \\
                       & score & $\%$ & score & $\%$ & score & $\%$ \\ 
\hline
QG$_{b}$ & 4.83 & \textbf{8.3} & 3.84 & 19.0 & 3.49 & 19.3 \\
QG$_{ours}$ & 4.83 & 7.3 & \textbf{3.89} & \textbf{24.7} & \textbf{3.58} & \textbf{23.3}  \\
\hline
\end{tabular}
\caption{Results of human evaluation for baseline and our best model. \textbf{score}: is the average score from raters; \textbf{\%}: represents the proportion of generated questions from one model that were rated higher than those from its counterpart. }
\label{tab:human-evaluation}
\end{table}
\vspace{-0.3cm}


\subsection{Synthetic Data Generation (RQ3)}

Further, can this QG framework be leveraged to generate effective synthetic data that can improve the closed-book QA task? Data augmentation is one of the main directions that question generation has been used for previously, with several studies finding improvements on the QA task~\cite{lewis-etal-2019-unsupervised,alberti-etal-2019-synthetic}. Here, we show how to leverage the proposed QG framework to improve closed-book QA tasks on seen data (WikiCQA) and unseen data (GooAQ and ELI5). Since freely available summary data is a good resource to generate synthetic data~\cite{lyu-etal-2021-improving}, we use WikiSum~\cite{cohen-etal-2021-wikisum}, which contains 39,775 coherent-paragraph summaries written by the article's authors on the WikiHow website. We take each sentence from the article summary as an answer and pass it into the best QG model, described in Section~\ref{subsec:qg}, to generate a question.
Then, we train an unsupervised QA model based on the synthetic QA pairs and optimize it by the following negative log-likelihood loss function:
\begin{equation}
    \mathcal{L}_{qa}=-\sum_{i=1}^{N}\log p(a_{i}|Q),
\end{equation}
where $a_{i}$ is the $i$-th token in the generated answer and Q is the question.

In this work, we pre-train a BART-base model on the 200K synthetic QA pairs that are generated through the best QG model (denoted as \textbf{QA}$_{s}$) and evaluate it on the test set of the seen dataset (WikiCQA) and unseen datasets (GooAQ-S and ELI5). This approach is unsupervised since the model is trained on no labeled question-answer pairs. The results are summarized in Table~\ref{tab:qa-results1}, showing 22.4\%, 12.5\%, and 11.4\% improvement than the BART-base model without any synthetic pre-training (denoted as \textbf{QA}$_{b}$), on WikiCQA, GooAQ-S, and ELI5, respectively. This shows that pre-training on the generated question-answer pairs derived from our QG model leads to significant improvements, echoing findings in previous works that find synthetic data can be helpful~\cite{khashabi-etal-2021-gooaq-open,lewis-etal-2021-paq,ding2021learning}. After further fine-tuning on the target training set (denoted as \textbf{QA}$_{s+f}$), from Table~\ref{tab:qa-results2} we can see that~\textbf{QA}$_{s+f}$ also achieves better results than fine-tuned baselines model \textbf{QA}$_{b+f}$ by 3.6\%, 0.4\%, and 4.3\% on the same three datasets.

\begin{table}[htbp!]
\begin{subtable}{\linewidth}
\centering\small
\begin{tabular}{p{1cm}ccccc}
\hline
\textbf{Dataset} & \textbf{Model} & \textbf{R-1} & \textbf{R-2} & \textbf{R-L} & \textbf{R-Lsum} \\
\hline
\multirow{2}{*}{WikiCQA} & QA$_{b}$ & 18.41 & 5.53 & 14.72 & 15.98 \\
                         & QA$_{s}$ & \textbf{24.10} & \textbf{7.07} & \textbf{18.02} & \textbf{20.31} \\ \hline
\multirow{2}{*}{GooAQ-S} & QA$_{b}$ & 18.58 & 5.80 & 14.77 & 15.66 \\
                       & QA$_{s}$ & \textbf{21.93} & \textbf{6.12} & \textbf{16.62}  & \textbf{18.35} \\ \hline
\multirow{2}{*}{ELI5} & QA$_{b}$ & 12.38 & 2.19 & 8.98 & 10.63 \\
                      & QA$_{s}$ & \textbf{13.97} & \textbf{2.29} & \textbf{10.00} & \textbf{11.77} \\  
\hline
\end{tabular}
\caption{Unsupervised QA results.}
\label{tab:qa-results1}
\end{subtable}

\begin{subtable}{\linewidth}
\centering\small
\begin{tabular}{p{1cm}ccccc}
\hline
\textbf{Dataset} & \textbf{Model} & \textbf{R-1} & \textbf{R-2} & \textbf{R-L} & \textbf{R-Lsum} \\
\hline
\multirow{2}{*}{WikiCQA} & QA$_{b+f}$ & 27.16 & 7.32 & 19.85 & 23.49 \\          
                         & QA$_{s+f}$ & \textbf{28.44} & \textbf{8.14} & \textbf{20.56} & \textbf{24.77} \\ \hline
\multirow{2}{*}{GooAQ-S} & QA$_{b+f}$ & 27.72 & 7.68 & 20.08 & 23.64\\
                       & QA$_{s+f}$ & \textbf{28.44} & \textbf{7.73} & \textbf{20.17} & \textbf{24.12}  \\ \hline
\multirow{2}{*}{ELI5} & QA$_{b+f}$ & 22.06 & 3.93 & 13.93 & 19.72 \\
                      & QA$_{s+f}$ & \textbf{23.37} & \textbf{4.27} & \textbf{14.53} & \textbf{20.88} \\
\hline
\end{tabular}
\caption{Semi-supervised QA results.}
\label{tab:qa-results2}
\end{subtable}
\caption{Evaluation on QA tasks. \textbf{QA$_{s/b}$}: QA model w/o pre-training on synthetic data; \textbf{QA$_{*+f}$}: QA$_{*}$ fine-tuned on target dataset.}
\label{tab:qa-results3}
\end{table}
\vspace{-0.5cm}
\subsection{Ablation Study (RQ4)}
Finally, we investigate the contribution of each component in the proposed closed-book QG framework. Table~\ref{tab:ablation-results} shows the results on the WikiCQA dataset. As described in Section~\ref{subsec:cl}, the  contrastive learning module takes ground-truth questions as positive pairs to the input answers (denoted as \textbf{CL}$_{t}$)
We also explore its variant -- in which we feed the same input (answers) to the encoder with different dropouts to obtain the positive pairs~\cite{gao-etal-2021-simcse} (denoted as \textbf{CL}$_{s}$). We find that the choice of \textbf{CL}$_{t}$ is slightly better than \textbf{CL}$_{s}$ in most evaluation metrics. In addition, by adding the answer reconstruction (\textbf{AR}) module, the model performance can be further improved. Thus, we can observe the effectiveness of the proposed contrastive learning module and the overall design of the framework.

\begin{table}[htbp!]
\centering\small
\begin{tabular}{lccccccc}
\hline
\textbf{Models} & \textbf{R-1} & \textbf{R-2} & \textbf{R-L} & \textbf{R-LSum}\\
\hline
QG$_{b}$ & 48.39 & 26.84 & 46.08 & 46.16  \\
QG$_{b}$+CL$_{s}$ & 48.67 & 27.34 & 46.43 & 46.46 \\
QG$_{b}$+CL$_{t}$ & 48.72 & 27.26 & 46.33 & 46.41 \\
QG$_{b}$+CL$_{s}$+AR & \textbf{49.26} & 27.77 & 46.82 & 46.88 \\
QG$_{b}$+CL$_{t}$+AR & 49.22 & \textbf{27.79} & \textbf{46.98} & \textbf{47.08} \\
\hline
\end{tabular}
\caption{Ablation study of our proposed QG framework on WikiCQA. \textbf{CL}$_*$: contrastive learning module; \textbf{AR}: answer reconstruction module.}
\label{tab:ablation-results}
\end{table}
\vspace{-0.5cm}




\section{Case Study}

Finally, we showcase some examples of the generated questions on the WikiCQA dataset. 
For the first example in Table~\ref{tab:good-examples}, we can see our model generates ``how many'' for the statistics-type answer, which is closer to the reference ``what percentage'' than the baseline model, generating a ``yes/no'' question. For the second one, although the reference contains the detailed information ``textured surface'' which is not in the answer, our model captures the information ``without nails'', which is more informative than the baseline model. Thus, the proposed model can produce better questions that are more relevant to the answers and contain more detailed information than questions generated from baseline models.
There are also some cases where the proposed model fails to generate good questions shown in Table~\ref{tab:bad-examples}. 
Compared with the reference question ``what food can be included on a clear liquid diet'', our model generates ``what can I eat to lose weight'', which fails in automatic evaluation metrics and is hard for annotators to justify in terms of relevance. 
For the second one, although the ``broken door'' generated from our model is relevant to the answer, it's still far from the reference ``door sticking to the weather strip''. The question generated from the baseline model related to ``bedroom'' is even less relevant. These examples encourage us to explore how to capture more semantics in the answers and generate questions that have more detailed information in the future.

\begin{table}
\centering
\begin{subtable}{\linewidth}
\centering\small
\begin{tabular}{p{0.01\textwidth}p{0.8\textwidth}}
\hline
A & The most recent statistics show that around 2.5\% of small businesses are audited by the irs. \\
Q$_{B}$ & Are small businesses audited by the irs? \\
Q$_{G}$ & How many small businesses are audited by the irs? \\
Q$_{R}$ & What percentage of small businesses are audited? \\ \hline
A & Try using poster putty to secure the images in your collage. if that doesn't work, you may have to use nails. \\
Q$_{B}$ & How do i put pictures in a collage? \\
Q$_{G}$ & How do i make a collage without nails? \\
Q$_{R}$ & How can i make a collage on a wall with a textured surface? \\
\hline
\end{tabular}
\caption{Good examples.}
\label{tab:good-examples}
\end{subtable}

\begin{subtable}{\linewidth}
\centering\small
\begin{tabular}{p{0.01\textwidth}p{0.8\textwidth}}
\hline
A & Foods such as popsicles, hard candy, and gelatin can be eaten on a clear liquid diet. \\
Q$_{B}$ & what can you eat on a clear liquid diet? \\
Q$_{G}$ & What can I eat to lose weight? \\
Q$_{R}$ & What food can be included on a clear liquid diet? \\ \hline
A & You'll have to remove the door and sand, prime, and repaint it. \\
Q$_{B}$ & What do i have to do to make my bedroom look nice? \\
Q$_{G}$ & How do i fix a broken door? \\
Q$_{R}$ & What do i do if my door is sticking to the weather strip? \\
\hline
\end{tabular}
\caption{Bad examples.}
\label{tab:bad-examples}
\end{subtable}
\caption{Generated question examples. \textbf{Q}$_{B}$: generated questions from baseline model; \textbf{Q}$_{G}$: generated questions from proposed QG model;
\textbf{Q}$_{R}$: reference questions.}
\vspace{-0.5cm}

\label{tab:examples}
\end{table}

\section{Conclusion}

In this work, aiming to empower a QG model to better understand the semantics of long-form abstractive answers and store more information in its parameters, we propose a closed-book QG model empowered by a contrastive learning module and an answer reconstruction module. We present a new \textit{closed-book} long-form QA dataset -- WikiCQA involving more than 20K real user QA pairs and show that WikiCQA is a valuable training resource, complementing public datasets.
Through the experiments, the proposed QG model shows better performance than baselines through automatic and human evaluations.
Moreover, we show how to leverage the proposed model as a data augmentation strategy to improve existing \textit{closed-book} QA systems.
The closed-book QA model, pre-trained on our generated synthetic QA pairs, achieves better performance on the seen dataset (WikiCQA). In addition, it shows strong generalization on unseen datasets (GooAQ and ELI5), which further demonstrates the effectiveness of our QG framework for enhancing closed-book QA performance. 


\section*{Limitations}
While our model shows promising results on English datasets, its efficiency and performance in other languages require further investigation. Moreover, our evaluation is limited to datasets with one-to-one QA pairs, without considering situations where multiple answers correspond to a single question.
Before feeding the input into the model, we ignore the tokens that exceed the maximum length we set, which may sometimes bring in an information loss for the input corpus.
Additionally, our model is built on transformer architecture and therefore requires huge computation resources, especially for large training data sizes.
Our focus in this paper is solely on the closed-book setting and our experiments center around four research questions to evaluate the effectiveness of the proposed framework, which we believe is an essential and emerging area deserving of specialized research investigation. In future work, we aim to explore the feasibility of adapting our framework for other QG scenarios.


\section*{Acknowledgements}

We thank Zhuoer Wang, Yun He, Ziwei Zhu, anonymous reviewers, and the action editor for providing valuable feedback.

\bibliography{anthology,custom}
\bibliographystyle{acl_natbib}

\appendix
\gdef\thesection{Appendix \Alph{section}}

\section{Data Filtering}
\label{sec:data-preprocess}

First, we choose the question-answer pairs where the questions start with a question-word in the list [`how', `what', `can', `is', `do', `why', `are', `does', `where', `when', `should', `will', `did', `which', `who', `would', `if', `about', `for', `as', `could', `in', `after', `at', `while', `to', `am', `has', `any'] and end with a question mark. Then, we filter out the pairs where the length of answers is less than 8 tokens or the answers are meaningless. For the ``meaningless'' answers we mean, answers are ``please refer to certain website http://xxx'' or ``please refer to the article xxx''.

\section{Dataset Analysis}
\label{sec:dataset-analysis}

\begin{figure*}[htbp!]
\centering
    \begin{subfigure}[!]{0.20\textwidth}
    \begin{tikzpicture}[scale=0.65]
\begin{axis}[
    width=1.75\linewidth,
    height=\axisdefaultheight,
    xtick=data,
	symbolic x coords={0-7, 8-15, 16-31, 32-63},
	ylabel=Percentage(\%),
	xticklabel style={rotate=30},
	ybar,
	bar width=4.5pt,
	ymajorgrids=true,
    grid style=dashed,
    axis lines*=left,
    ymin=0.0, ymax=80,
    nodes near coords,
    nodes near coords align={vertical},
    every axis plot/.append style={fill},
    cycle list/Paired,
	]
\addplot
    coordinates{(0-7,13) (8-15,72) (16-31,13) (32-63, 2)};
\addplot
    coordinates{(0-7,30) (8-15,70) (16-31,1) (32-63, 0)};
\addplot
    coordinates{(0-7,12) (8-15,40) (16-31,40) (32-63, 7)};
\legend{WikiCQA, GooAQ-S, ELI5}
\end{axis}
\end{tikzpicture}
    \vspace{-0.5cm}
    \caption{Question length.}
    \label{fig:question-length}
    \end{subfigure}
    \hspace{\fill}
    \begin{subfigure}[!]{0.30\textwidth}
    \begin{tikzpicture}[scale=0.65]
\begin{axis}[
    width=1.50\linewidth,
    height=\axisdefaultheight,
    xtick=data,
	symbolic x coords={8-15, 16-31, 32-63,  64-127, 128-255, 256-511},
	xticklabel style={rotate=30},
	ylabel=Percentage(\%),
	ybar,
	bar width=4.pt,
    axis lines*=left,
	ymin=0.0, ymax=70,
	ymajorgrids=true,
    grid style=dashed,
    nodes near coords,
    nodes near coords align={vertical},
    every axis plot/.append style={fill},
    cycle list/Paired,
	]
\addplot
    coordinates{(8-15,1) (16-31,25) (32-63, 44) (64-127, 27) (128-255, 3) (256-511, 0)};
\addplot
    coordinates{(8-15,1) (16-31,8) (32-63, 66) (64-127, 25) (128-255, 0) (256-511, 0)};
\addplot
    coordinates{(8-15,1) (16-31,8) (32-63, 22) (64-127, 30) (128-255, 24) (256-511, 15)};
\legend{WikiCQA, GooAQ-S, ELI5}
\end{axis}
\end{tikzpicture}
    \vspace{-0.5cm}
    \caption{Answer length.}
    \label{fig:answer-length}
    \end{subfigure}
    \hspace{\fill}
    \begin{subfigure}[!]{0.45\textwidth}
    \begin{tikzpicture}[scale=0.65]
\begin{axis}[
    width=1.40\linewidth,
    height=\axisdefaultheight,
    xtick=data,
    x tick label style={
		/pgf/number format/1000 sep=},
	symbolic x coords={why, how, what, if, when, is, can, do, where, are, which},
	xticklabel style={rotate=30},
	ylabel=Percentage(\%),
	ybar,
	axis lines*=left,
	ymin=0.0, ymax=40,
	bar width=3pt,
	ymajorgrids=true,
    grid style=dashed,
    nodes near coords,
    nodes near coords align={vertical},
    every axis plot/.append style={fill},
    cycle list/Paired,
	]
\addplot
    coordinates{(why,2)(how,36)(what,22)(if,2)(when,1)(is,6)(can,16)(do,5)(where,1)(are,2)};
\addplot
    coordinates{(why,7)(how,7)(what,10)(if,1)(when,1)(is,18)(can,9)(do,6)(where,2)(are,23)};
\addplot
    coordinates{(why,38)(how,21)(what,14)(if,4)(when,2)(is,1)(can,1)(do,1)(where,1)};
\legend{WikiCQA, GooAQ-S, ELI5}
\end{axis}
\end{tikzpicture}
    \vspace{-0.5cm}
    \caption{Common question-word.}
    \label{fig:question-unigram}
    \end{subfigure}
    \hspace{\fill}
 \caption{Comparison of the distribution of QA length and question types among three datasets.}    
\label{fig:data-analysis}
\end{figure*}
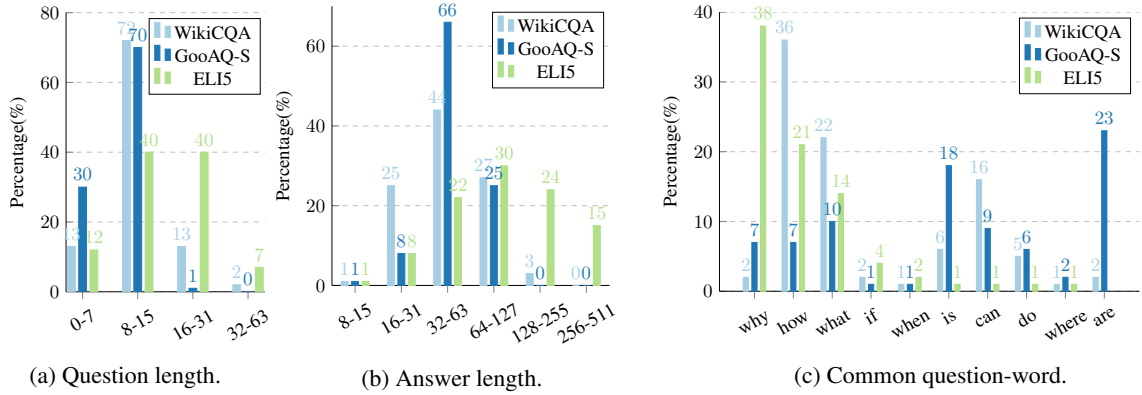

To better understand the content of WikiCQA in comparison with existing QA datasets, Figure~\ref{fig:data-analysis} shows the distributions of question length, answer length, and common question word. The questions and answers are tokenized by BART~\cite{lewis-etal-2020-bart}. From Figure~\ref{fig:question-length} and ~\ref{fig:answer-length} we could see more than 80\% of the questions in WikiCQA and GooAQ-S have less than 15 tokens and more than 90\% of answers of them lie in the range of 16-127 tokens while both questions and answers in ELI5 have a broader range of lengths. By measuring the proportion of the leading unigram that starts a question (Figure~\ref{fig:question-unigram}), the dominant pattern is ``how'' questions in WikiCQA, while ELI5 has many ``why'' questions. We think WikiCQA has the potential to benefit the community in three ways. Firstly, its source is unique compared to the current two closed-book data sources, as it is collected from the Q\&A section of articles on WikiHow. The questions and answers are related to the specific articles, asked by WikiHow users who read the articles and answered by a knowledgeable reader or editor. Secondly, compared to large-scale QA datasets, WikiCQA has longer answers and more open-ended questions with a wider range of question types. Additionally, it has a different data distribution compared to the other two datasets, which is crucial for testing the generalization ability of question-answering models. Finally, the question-answer pairs within relevant articles in WikiCQA can also be used as a valuable resource for conversational question-answering tasks. Our code and data can be found at \href{https://github.com/dongxiangjue/Closed-book-Question-Generation-via-Contrastive-Learning}{https://github.com/dongxiangjue/Closed-book-Question-Generation-via-Contrastive-Learning}.

\end{document}